\documentclass{article}
\usepackage{iclr2027_conference,times}

\usepackage[utf8]{inputenc}
\usepackage[T1]{fontenc}
\usepackage{hyperref}
\usepackage{url}
\usepackage{booktabs}
\usepackage{multirow}
\usepackage{amsfonts}
\usepackage{nicefrac}
\usepackage{microtype}
\usepackage{xcolor}
\usepackage{graphicx}
\usepackage{pifont}
\usepackage{subcaption}
\usepackage{caption}
\usepackage{adjustbox}
\usepackage{amsmath}
\usepackage{amssymb}
\usepackage{mathtools}
\usepackage{float}
\usepackage{needspace}
\usepackage{placeins}
\hypersetup{hidelinks}

\usepackage{graphicx}
\usepackage{enumitem}

\title{PP-CFG: Polarity-Preserving Guidance\\for Controllable Few-Step\\Image and Video Generation}

\author{%
  \textbf{Junyi He\textsuperscript{1*}, Xu Peng\textsuperscript{2*}, Chengming Xu\textsuperscript{2*}, Sixiao Zheng\textsuperscript{1*}, Donghao Luo\textsuperscript{2}}\\[2pt]
  \textbf{Chao Wen\textsuperscript{2}, Peng TANG\textsuperscript{2}, Xijie Huang\textsuperscript{1}, Chengjie Wang\textsuperscript{2}, Yanwei Fu\textsuperscript{1}}\\[4pt]
  \normalfont\small \textsuperscript{1}Fudan University\qquad \textsuperscript{2}Tencent YouTu Lab\\
  \normalfont\small * Equal contribution}
\iclrfinalcopy
\hypersetup{
  pdftitle={PP-CFG: Polarity-Preserving Guidance for Controllable Few-Step Image and Video Generation},
  pdfauthor={Junyi He, Xu Peng, Chengming Xu, Sixiao Zheng, Donghao Luo, Chao Wen, Peng TANG, Xijie Huang, Chengjie Wang, Yanwei Fu}
}

\begin{document}
\raggedbottom
\maketitle
\lhead{Preprint}

\begin{abstract}
Few-step diffusion distillation reduces sampling cost but often weakens adjustable classifier-free guidance (CFG) and runtime negative prompting. We introduce PP-CFG, an architecture that keeps positive and negative text conditions separate and combines their attention responses with opposite signs within a shared visual backbone. Its Scale-Time Polarity Mixer refines each condition using the guidance scale and denoising timestep, allowing prompt influence to vary throughout generation. The student retains a continuous scale input and selectable negative prompts with one visual backbone pass per step. We evaluate PP-CFG on Wan2.1 for video generation with distribution matching distillation (DMD), and on Qwen-Image with both latent consistency model (LCM) distillation and DMD. PP-CFG achieves 81.46 VBench in four steps and lower video teacher--student distances across guidance scales than step-distilled DMD. On NegElement-1K, our benchmark of 1,000 matched image and video prompt cases, it achieves higher target-removal rates than the corresponding teachers. These experiments support the method's generality across attention architectures, generation modalities, and distillation objectives while retaining competitive generation quality.
\end{abstract}

\section{Introduction}

Diffusion and flow-matching models produce high-quality images and videos~\citep{wan2025wan,wu2025qwenimage}, but repeated denoising evaluations make generation expensive. Few-step distillation reduces this cost by transferring a teacher's generation behavior to a student~\citep{yin2024one,yin2024improved}. Users also need to adjust prompt adherence and suppress unwanted content as their prompts and preferences change. A practical student should retain both controls after distillation.

Classifier-free guidance (CFG) combines positive and negative predictions to control generation~\citep{ho2022classifier}. Fixed-guidance distillation absorbs guided predictions into student weights, typically losing runtime scale adjustment and negative prompting. Applying external CFG after distillation can degrade output quality (Fig.~\ref{fig:teaser}). Conditioning the student on guidance scale restores an adjustable input~\citep{meng2023distillation}, but does not by itself retain a selectable negative prompt. Attention-based guidance provides a negative-prompt path~\citep{chen2025normalized,guo2025vsf}. Table~\ref{tab:method_comparison} summarizes the considered configurations. The challenge is to retain both controls within one visual backbone pass while preserving the distinct roles of the two prompts.

\newsavebox{\capabilitytable}
\newlength{\capabilityheight}
\sbox{\capabilitytable}{%
  \small\setlength{\tabcolsep}{2.5pt}\renewcommand{\arraystretch}{1.3}%
  \begin{tabular}{@{}lccc@{}}
    \toprule
    \textbf{Setting} & \textbf{Adj.} & \textbf{Neg.} & \textbf{1-NFE} \\
    \midrule
    Fixed guidance & \ding{55} & \ding{55} & \ding{51} \\
    LCM (SD) & \ding{51} & \ding{55} & \ding{55} \\
    DMD (SD) & \ding{51} & \ding{55} & \ding{55} \\
    LCM (CFG-E+SD) & \ding{51} & \ding{55} & \ding{51} \\
    DMD (CFG-E+SD) & \ding{51} & \ding{55} & \ding{51} \\
    PCM & \ding{51} & \ding{51} & \ding{55} \\
    \textbf{PP-CFG} & \ding{51} & \ding{51} & \ding{51} \\
    \bottomrule
  \end{tabular}}
\setlength{\capabilityheight}{\dimexpr\ht\capabilitytable+\dp\capabilitytable\relax}

\begin{figure}[t]
  \centering
  \noindent\begin{minipage}[t]{0.485\linewidth}
    \vspace{0pt}\centering\usebox{\capabilitytable}
  \end{minipage}\hfill
  \begin{minipage}[t]{0.485\linewidth}
    \vspace{0pt}\centering
    \includegraphics[
  width=\linewidth,
  height=\capabilityheight,
  keepaspectratio
]{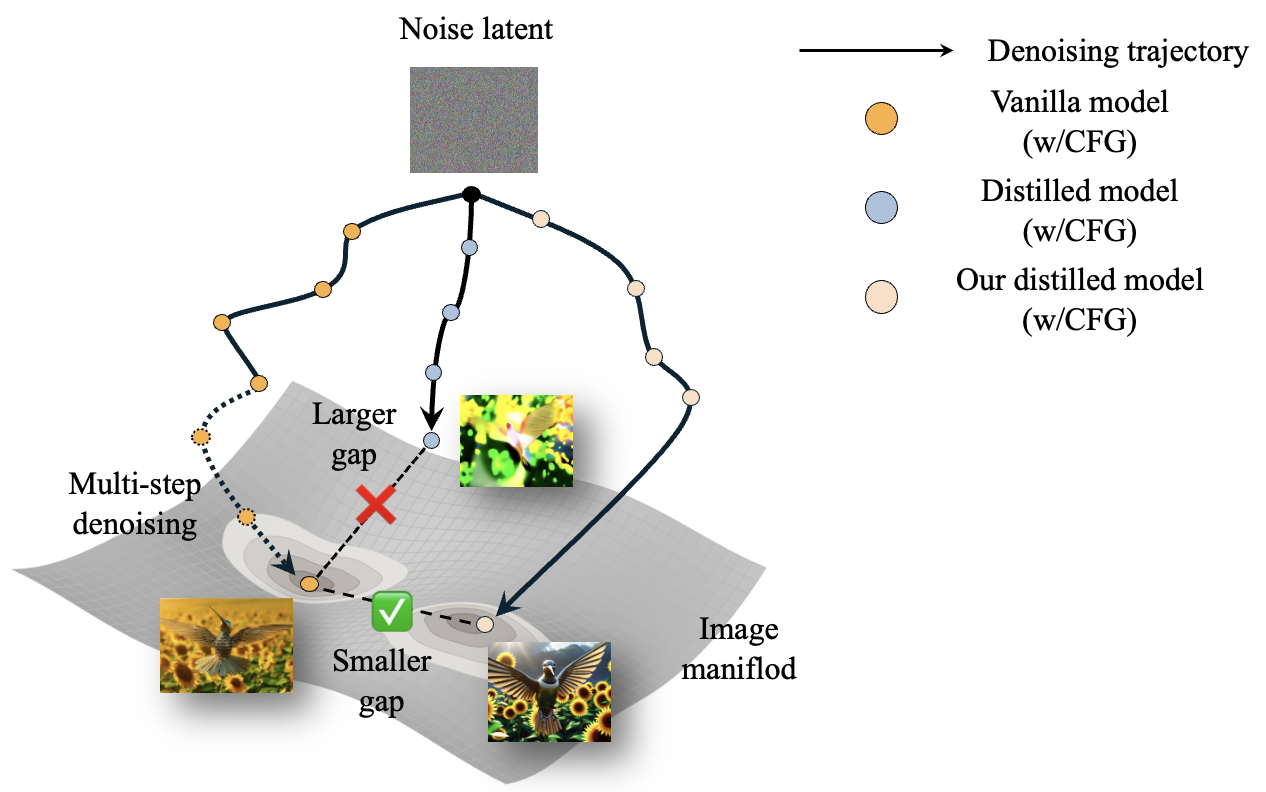}
  \end{minipage}\par\vspace{6pt}
  \noindent\begin{minipage}[t]{0.485\linewidth}
    \vspace{0pt}\captionsetup{position=bottom,skip=0pt}
    \captionof{table}{Runtime controls. Adj.: adjustable scale; Neg.: negative prompts; \mbox{1-NFE}: one visual backbone pass per step. SD: step distillation; \mbox{CFG-E}: CFG embedding; PCM: phased consistency model.}
    \label{tab:method_comparison}
  \end{minipage}\hfill
  \begin{minipage}[t]{0.485\linewidth}
    \vspace{0pt}\captionsetup{position=bottom,skip=0pt}
    \captionof{figure}{\textbf{CFG after distillation.} External CFG can degrade student outputs and increase their discrepancy from the teacher. PP-CFG learns scale-dependent guidance within a shared visual stream.}
    \label{fig:teaser}
  \end{minipage}
\end{figure}

We propose Polarity-Preserving CFG (PP-CFG), where \emph{polarity} denotes the positive and negative roles of the text conditions. The method keeps these contexts separate and uses a Scale-Time Polarity Mixer to refine each one according to the requested guidance scale and denoising timestep. These refinements adjust how visual features attend to each prompt and what information the text provides at each denoising stage. Following Negative-Away Steer Attention (NASA)~\citep{nguyen2025supercharged}, polarity blocks subtract the negative attention response from the positive response to update one shared visual stream (Fig.~\ref{fig:framework}). We jointly train the mixer and backbone using either LCM or DMD. At inference, users can choose both the guidance scale and negative prompt while the model uses one visual backbone pass per step.

\Needspace{4\baselineskip}
We evaluate guidance adjustment and generation quality, and introduce NegElement-1K to assess whether models suppress specified elements through negative prompting. The benchmark uses matched image and video prompt cases to examine this capability in both modalities. Our experiments show that PP-CFG retains {both adjustable CFG and effective target suppression after few-step distillation} while maintaining competitive generation quality. We summarize our contributions as follows:
\begin{itemize}
  \item We introduce PP-CFG, which {combines our Scale-Time Polarity Mixer for adjustable CFG with NASA-style attention subtraction for target-specific suppression} in one visual backbone pass {per denoising step}.
  \item We introduce NegElement-1K, a benchmark of 1,000 matched image and video prompt cases for evaluating target removal through negative prompting.
  \item We evaluate PP-CFG with DMD on Wan and with both LCM and DMD on Qwen-Image. These experiments support its generality across attention architectures, generation modalities, and distillation objectives.
\end{itemize}

\section{Related Work}

\noindent\textbf{Diffusion distillation.}
Few-step distillation accelerates image and video diffusion models~\citep{luo2023latent,yin2024improved,wang2023videolcm}. For video, DCM separates semantic and detail modeling~\citep{lv2025dcm}, while Self-Forcing trains on inference rollouts~\citep{huang2025self}. CausVid learns causal generators from bidirectional teachers~\citep{yin2025slow}; TMD matches teacher ODE transition kernels~\citep{nie2026transition}. ASD aligns generations across denoising step counts through adversarial self-distillation~\citep{yang2025towards}. AVD uses adaptive regression and temporal regularization to mitigate oversaturation and temporal collapse~\citep{you2026adaptivevideo}. These methods improve sampling efficiency and generation quality. Our focus is preserving guidance-scale and negative-prompt controls after distillation.

\noindent\textbf{Guidance and negative prompting.}
CFG combines conditional and unconditional predictions using two model evaluations~\citep{ho2022classifier}. Guidance distillation replaces these evaluations with a scale-conditioned student~\citep{meng2023distillation}. P-Guide steers the initial latent prior~\citep{peng2026p}, while Internal Guidance combines intermediate and final transformer outputs~\citep{zhou2026guiding}. Other work improves guidance through manifold constraints~\citep{chung2024cfg++}, projection~\citep{sadat2024eliminating}, or restricted intervals~\citep{kynkaanniemi2024applying}. For attention-space negative prompting, NASA subtracts responses to negative concepts~\citep{nguyen2025supercharged}, and Normalized Attention Guidance (NAG) controls the resulting feature magnitude~\citep{chen2025normalized}. PP-CFG combines response subtraction with separate text refinements conditioned on scale and time, retaining both guidance-scale and negative-prompt inputs.

\section{Method}
\label{sec:method}

\begin{figure}[t]
  \centering
  \includegraphics[width=0.97\linewidth]{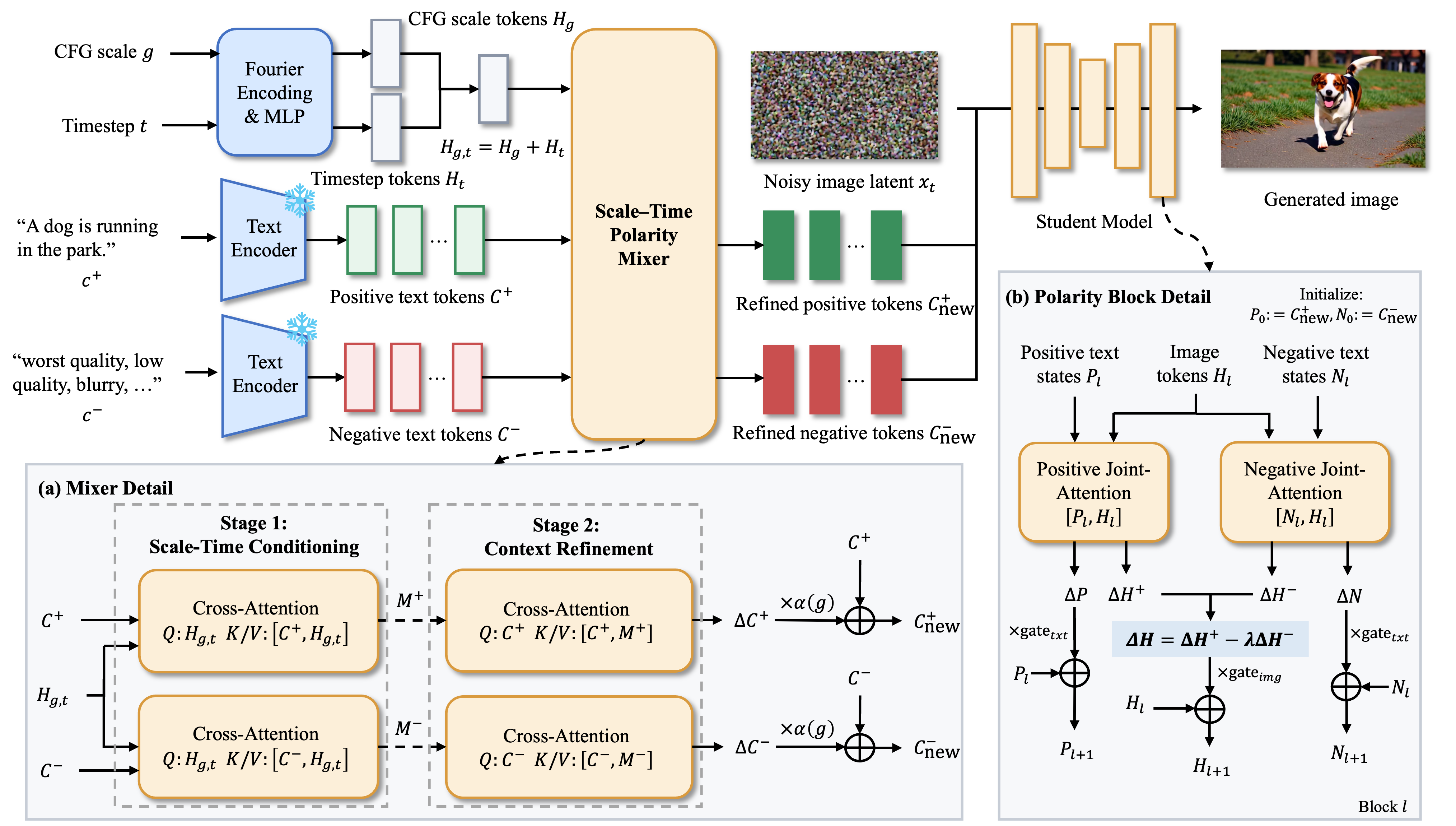}
  \caption{\textbf{PP-CFG on Qwen-Image.} (a) The mixer refines each text condition using guidance scale $g$ and denoising time $t$; internal projections are omitted in the diagram. (b) Joint-attention paths update separate text states and combine their visual responses within one shared stream. Section~\ref{sec:polarity-blocks} describes this instance and the Wan cross-attention implementation.}
  \label{fig:framework}
\end{figure}

\subsection{Guidance Formulation and Overview}

At each denoising step, the model receives a noisy latent $x_t$ at time $t$, a positive prompt $c^+$ describing desired content, a negative prompt $c^-$ naming unwanted content (or an empty prompt), and guidance scale $g$. Let $v_\theta(x_t,t,c)$ denote a backbone prediction under prompt $c$. Standard CFG combines two such predictions:
\begin{equation}
  v_{\mathrm{cfg}}(x_t,t,c^+,c^-;g)
  =v_\theta(x_t,t,c^-)
  +g\bigl(v_\theta(x_t,t,c^+)-v_\theta(x_t,t,c^-)\bigr).
  \label{eq:cfg}
\end{equation}
The two evaluations let visual features evolve separately under each prompt. PP-CFG maintains one visual state: its mixer refines the two text conditions using $(g,t)$, and polarity blocks combine their attention responses. This retains a separate negative-prompt input without two full backbone evaluations.

At $g=1$, Eq.~\ref{eq:cfg} reduces to the positive prediction. PP-CFG does not apply this equation at its output, so its endpoint behavior depends on the mixer and attention combination defined below.

\subsection{Scale-Time Polarity Mixer}

The mixer learns to adapt each text condition while the two attention paths share visual features. In attention from visual to text tokens, visual queries interact with text keys to determine attention weights, which are used to aggregate text values. Transforming the keys can express some matching variations otherwise produced by branch-specific query transformations; transforming the values also changes the information delivered to visual tokens. Conditioning these refinements on $(g,t)$ allows each prompt's influence to vary with guidance strength and denoising stage.

A frozen text encoder maps $c^\pm$ to token sequences $C^\pm$. Sinusoidal embeddings followed by multilayer perceptrons, denoted $E_g$ and $E_t$, produce scale and time tokens with matching lengths and widths:
\begin{equation}
  H_g=E_g(g),\qquad H_t=E_t(t),\qquad H_{g,t}=H_g+H_t.
  \label{eq:scale-time-tokens}
\end{equation}
Their elementwise sum supplies both controls without increasing the number of conditioning tokens.

Two attention stages connect these controls to each prompt. Let $X^\pm$ and $S$ be projections of $C^\pm$ and $H_{g,t}$ into a common mixer feature space; $[A,B]$ denotes concatenation along the token dimension. Each stage $F_i(Q,K\!V)$ applies attention followed by a feed-forward transformation, with its first input supplying queries and its second supplying keys and values:
\begin{equation}
  M^\pm=F_1\bigl(S,[X^\pm,S]\bigr),\qquad
  U^\pm=F_2\bigl(X^\pm,[X^\pm,M^\pm]\bigr).
  \label{eq:polarity-mixer-stages}
\end{equation}
Stage~1 lets scale-time tokens read the corresponding prompt, producing a prompt-specific control representation $M^\pm$. Stage~2 passes that information back to the text tokens, producing $U^\pm$. The two stages have distinct parameters, each shared across polarities; neither path reads the other prompt. This preserves separate contexts before their visual responses are combined.

Separate output projections $W^+$ and $W^-$ map the refined features back to the text-context width. The gate $\alpha(g)$ controls the strength of this residual update:
\begin{equation}
  \Delta C^\pm=W^\pm U^\pm,\qquad
  C_{\mathrm{new}}^\pm=C^\pm+\alpha(g)\Delta C^\pm.
  \label{eq:polarity-mixer-update}
\end{equation}
The scale tokens influence the content of the update, whereas $\alpha(g)$ scales its contribution. Distinct projections allow the two prompt roles to receive different updates despite sharing the attention stages. Initializing $W^\pm$ to zero makes the refined contexts equal to the original text representations at initialization. We use a square-root gate $\alpha(g)=\sqrt{(g-g_{\min})/(g_{\max}-g_{\min})}$, where $g_{\min}=1$ and the configuration-specific $g_{\max}$ is reported in Table~\ref{tab:implementation}. This concave mapping makes the gate more responsive at low guidance scales while reducing its sensitivity as $g$ increases.

\subsection{Polarity Blocks across Backbones}
\label{sec:polarity-blocks}

Each polarity block reads the refined contexts from a shared visual state $H_l$, where $l$ indexes backbone blocks. Let $\Delta H_l^+$ and $\Delta H_l^-$ denote the visual attention responses under the positive and negative conditions. Following NASA~\citep{nguyen2025supercharged}, the block combines them before updating the visual stream:
\begin{equation}
  \Delta H_l=\Delta H_l^+-\lambda\Delta H_l^-.
  \label{eq:polarity-combination}
\end{equation}
Here $\lambda$ weights the negative attention response; it has a different role from the runtime scale $g$ supplied to the mixer. Subtraction acts on intermediate attention features, with the final guided prediction learned during distillation. The polarity mechanism is enabled by default with a fixed, configuration-specific $\lambda$ reported in Table~\ref{tab:implementation}, but can be disabled at $g=1$.

\noindent\textbf{Wan cross-attention.}
Visual tokens supply shared queries, while $C_{\mathrm{new}}^+$ and $C_{\mathrm{new}}^-$ supply separate text keys and values. The two calculations share projection weights and are packed along the batch dimension. The text contexts are not updated by the backbone's cross-attention; their visual responses are combined using Eq.~\ref{eq:polarity-combination}.

\noindent\textbf{Qwen-Image joint attention.}
As illustrated in Fig.~\ref{fig:framework}(b), initialize text states $P_0=C_{\mathrm{new}}^+$ and $N_0=C_{\mathrm{new}}^-$. At block $l$, separate joint-attention paths process $[P_l,H_l]$ and $[N_l,H_l]$. They update their text states independently and combine only their visual responses. The diagram's $\mathrm{gate}_{\mathrm{txt}}$ and $\mathrm{gate}_{\mathrm{img}}$ are the backbone's residual multipliers for text and image updates. Both implementations thus retain two prompt paths while advancing one visual stream.

\subsection{Training and Inference}
\label{sec:training}

PP-CFG can be trained with either latent consistency model (LCM) distillation~\citep{luo2023latent} or DMD~\citep{yin2024one,yin2024improved}. LCM encourages consistent predictions at latent states linked by a teacher denoising trajectory; DMD matches the student's generated distribution to the guided teacher distribution using teacher and critic score estimates. We use DMD for Wan and evaluate both objectives on Qwen-Image, jointly training the student diffusion transformer (DiT) and mixer while keeping the text encoder frozen.

For multi-guidance DMD, we sample $g$ uniformly from $[g_{\min},g_{\max}]$ and collect the conditions as $\eta_g=(c^+,c^-,g)$. Let $G_\phi(z,\eta_g)$ denote the complete few-step student sampler with parameters $\phi$ and initial noise $z$. Its generated clean latent is $x_0^\phi=G_\phi(z,\eta_g)$. DMD perturbs this output at a sampled noise level $\tau$ to obtain $y_\tau=a_\tau x_0^\phi+\sigma_\tau\epsilon$, where $\epsilon$ is independent standard Gaussian noise and $a_\tau,\sigma_\tau$ are the noising coefficients. Here $\tau$ is the time used to evaluate the distillation loss, whereas $t$ is the student's denoising time; $a_\tau$ is unrelated to the mixer gate $\alpha(g)$.

Denote the noised student and guided teacher distributions by $q_\tau^\phi(\cdot\mid\eta_g)$ and $p_\tau^{\mathrm{T}}(\cdot\mid\eta_g)$. At a given guidance setting, DMD aims to reduce their Kullback--Leibler divergence $D_{\mathrm{KL}}$:
\begin{equation}
  \mathcal{L}_{\mathrm{DMD}}(\phi;\eta_g)
  =\mathbb{E}_{\tau}\left[
    D_{\mathrm{KL}}\left(q_\tau^\phi(\cdot\mid\eta_g)\;\|\;p_\tau^{\mathrm{T}}(\cdot\mid\eta_g)\right)
  \right].
  \label{eq:dmd-objective}
\end{equation}
Training over a range of guidance strengths averages this objective over the sampled scales:
\begin{equation}
  \mathcal{L}_{\mathrm{MG\mbox{-}DMD}}(\phi)
  =\mathbb{E}_{g\sim\mathcal{U}(g_{\min},g_{\max})}
  \left[\mathcal{L}_{\mathrm{DMD}}(\phi;\eta_g)\right].
  \label{eq:mgdmd-objective}
\end{equation}

DMD optimizes this objective using a difference of scores, where a score is the gradient of a log density with respect to the noised latent $y_\tau$. The teacher provides a score estimate $s_{\mathrm{T}}$, while a critic with parameters $\psi$ estimates the student score $s_\psi$. The Jacobian $J_\phi(y_\tau)=\partial y_\tau/\partial\phi$ maps this signal to student parameters, and $w(\tau)$ weights noise levels. The student update is
\begin{equation}
  \nabla_\phi\mathcal{L}_{\mathrm{MG\mbox{-}DMD}}
  \approx\mathbb{E}_{z,\tau,\epsilon,\eta_g}\left[
    w(\tau)J_\phi(y_\tau)^\top
    \bigl(s_\psi(y_\tau,\tau,\eta_g)-s_{\mathrm{T}}(y_\tau,\tau,\eta_g)\bigr)
  \right].
  \label{eq:dmd-gradient}
\end{equation}
The teacher and critic use the same $g$ for their guided predictions, and the student receives it through the mixer. Appendix~\ref{app:implementation} gives the LCM objective and collects training configurations, including DMD weighting and the critic schedule.

At inference, each step refines $C^\pm$ using $(g,t)$ and applies the polarity blocks to $x_t$. One visual backbone pass produces the student prediction, avoiding two complete output-space CFG evaluations. The mixer and dual attention paths still add computation beyond an unguided pass.

\section{Experiments}
\subsection{Experimental Setup}
\noindent\textbf{Models and evaluation.}
We evaluate three combinations: Wan2.1-T2V-1.3B with DMD~\citep{wan2025wan}, and Qwen-Image with LCM or DMD~\citep{wu2025qwenimage}. VBench measures video quality and semantic alignment, reported as Total, Quality, and Semantic scores~\citep{huang2024vbench}. For images, GenEval tests compositional properties such as object count, color, and position~\citep{ghosh2023geneval}; DPG-Bench evaluates adherence to detailed prompts~\citep{hu2024ella}. Table~\ref{tab:main_results} gives sampling steps. Appendix~\ref{app:implementation} collects training configurations, and Appendix~\ref{app:evaluation} collects evaluation settings.

\noindent\textbf{Baselines.}
Video teacher and DMD, LCM, PCM, DCM, rCM, and AVD scores are from~\citet{you2026adaptivevideo}. We run ASD~\citep{yang2025towards} with its original configuration except for a chunk size of 1. We evaluate image students in our setup, including LCM and DMD variants, sCM~\citep{lu2024simplifying}, and MeanFlow~\citep{geng2025meanflows}. SD denotes step distillation with external CFG; CFG-E+SD additionally embeds the guidance scale in the student. These configurations distinguish external guidance from a learned scale input. Our image variants are labeled Ours (LCM) and Ours (DMD); unless stated otherwise, subsequent image results use Ours (LCM). The step-distilled DMD in the video scale sweep is labeled DMD (Step Distill) in Fig.~\ref{fig:cfg_adjustability}(b).

\noindent\textbf{NegElement-1K.}
\label{sec:negelement-1k}
NegElement-1K contains 1,000 template-generated cases balanced across ten categories (100 each). Each case pairs an image prompt containing a target element with a negative prompt naming it, a video prompt for the same scene, and a target-free reference prompt. Annotations identify the target, three scene elements to preserve, and target position, size, and difficulty. We generate four images or one video per case. A Qwen3.5-VL judge checks target presence in each image or sampled video frames. \emph{Removal success} is the fraction of samples judged target absent; preservation annotations do not enter this score. Failed or unparseable judgments count as failures. Appendix~\ref{app:negative-protocol} collects the benchmark and judge protocol.

\subsection{Main Quality Comparison}

\begin{table}[H]
  \centering
  \caption{Main generation results (0--100; higher is better). Bold marks the best listed student per metric. Teachers use two CFG evaluations per step. SD: step distillation; CFG-E: CFG embedding. The image teacher scores are reported by~\citet{wu2025qwenimage}; RL denotes their reinforcement-learning-enhanced model. Video result sources are specified above.}
  \label{tab:main_results}
  \small\setlength{\tabcolsep}{2.5pt}\renewcommand{\arraystretch}{1.08}
  \begin{minipage}[t]{0.45\linewidth}
    \vspace{0pt}
    \begin{tabular*}{\linewidth}{@{\extracolsep{\fill}}lrrrr@{}}
      \multicolumn{5}{c}{\textbf{Video: Wan2.1-T2V-1.3B}}\\[3pt]
      \toprule
      Method & Steps & Total & Quality & Semantic\\
      \midrule
      Teacher\vphantom{$91.00^{\mathrm{RL}}$} & 50 & 80.13 & 84.11 & 64.20\\
      \midrule
      DMD & 4 & 80.66 & \textbf{85.55} & 61.10\\[2.7pt]
      LCM & 6 & 72.12 & 77.62 & 50.15\\[2.7pt]
      PCM & 6 & 73.70 & 79.36 & 51.40\\[2.7pt]
      DCM & 6 & 73.92 & 79.27 & 52.16\\[2.7pt]
      rCM & 4 & 80.15 & 84.80 & 61.55\\[2.7pt]
      AVD & 4 & 81.35 & 85.27 & 65.65\\[2.7pt]
      ASD & 4 & 80.79 & 81.01 & \textbf{79.92}\\[2.7pt]
      \midrule
      \textbf{Ours} & 4 & \textbf{81.46} & 84.69 & 68.53\\[2.7pt]
      \bottomrule
    \end{tabular*}
  \end{minipage}\hfill
  \begin{minipage}[t]{0.52\linewidth}
    \vspace{0pt}
    \begin{tabular*}{\linewidth}{@{\extracolsep{\fill}}lrrr@{}}
      \multicolumn{4}{c}{\textbf{Image: Qwen-Image}}\\[3pt]
      \toprule
      Method & Steps & GenEval & DPG\\
      \midrule
      Teacher & 50 & $87.00/91.00^{\mathrm{RL}}$ & 88.32\\
      \midrule
      sCM & 8 & 60.00 & 85.54\\
      LCM & 8 & 61.85 & 80.69\\
      LCM (SD) & 8 & 81.94 & 87.02\\
      LCM (CFG-E+SD) & 8 & 81.64 & 86.90\\
      DMD & 8 & 79.08 & 86.40\\
      DMD (SD) & 8 & 80.72 & 86.71\\
      DMD (CFG-E+SD) & 8 & 80.59 & 86.68\\
      MeanFlow & 8 & 49.00 & 83.81\\
      \midrule
      \textbf{Ours (LCM)} & 8 & \textbf{82.39} & \textbf{87.05}\\
      \textbf{Ours (DMD)} & 8 & 80.22 & 86.98\\
      \bottomrule
    \end{tabular*}
  \end{minipage}
\end{table}

Table~\ref{tab:main_results} shows that PP-CFG retains competitive generation quality. On video, it achieves the highest listed student VBench Total (81.46) in four steps, versus 80.66 for DMD, 81.35 for AVD, and 80.13 for the teacher. DMD leads Quality and ASD leads Semantic, so the gains are not uniform across submetrics. On images, Ours (LCM) obtains the best student GenEval and DPG-Bench scores, modestly above LCM (SD). Ours (DMD) improves DPG-Bench over DMD (SD) but not GenEval; the teacher remains higher on both image metrics.

\begin{figure}[H]
  \centering
  \includegraphics[width=\linewidth]{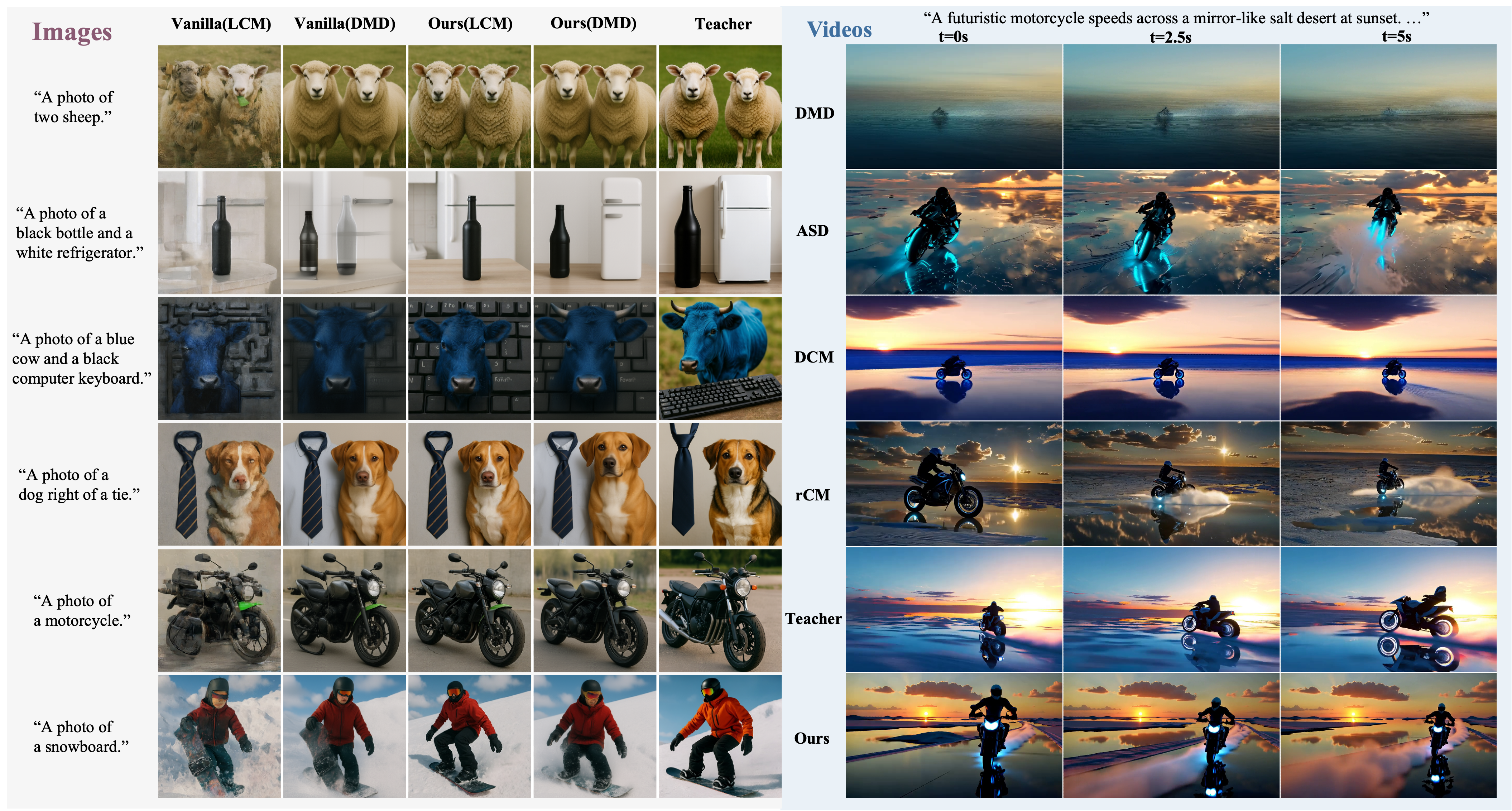}
  \caption{\textbf{Image and video generation quality.} Images compare vanilla LCM/DMD, their PP-CFG counterparts, and the Qwen-Image teacher under matched prompts. Videos compare PP-CFG, representative distillation baselines, and Wan2.1-T2V-1.3B, with three frames per video.}
  \label{fig:qualitative_comparison}
\end{figure}

Figure~\ref{fig:qualitative_comparison} complements these metrics with matched-prompt examples. PP-CFG retains clear object attributes in the images and coherent motorcycle and scene structure across the sampled video frames.

\begin{figure}[H]
  \centering
  \includegraphics[width=\linewidth]{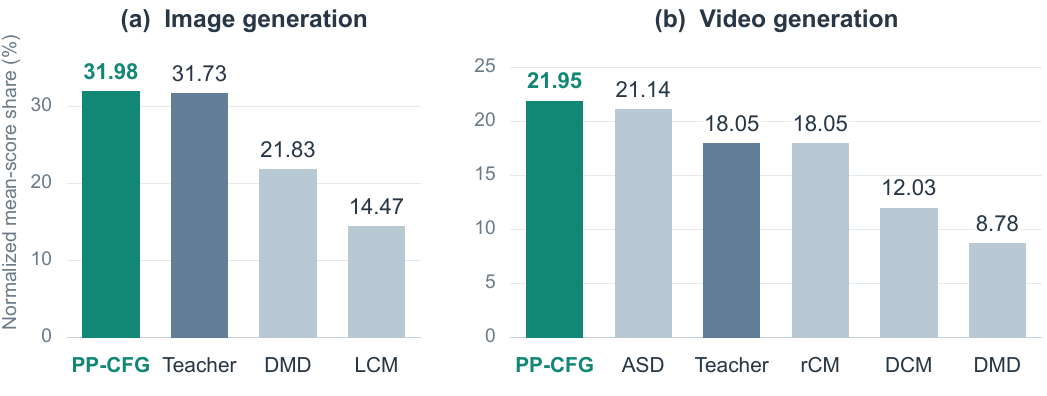}
  \caption{\textbf{User study.} Bars show each method's mean score divided by the sum of means within its modality (higher is better). PP-CFG is green; panels use different y-axis ranges.}
  \label{fig:user_study}
\end{figure}

\noindent\textbf{User study.}
\label{sec:user-study}
Ten participants evaluated visual quality and prompt alignment on a shared set of six video prompts and five image prompts. For video, each participant rated all six methods on every prompt using a 1--3 Likert scale (1 = poor, 3 = good); for image, each participant rated all four methods on every prompt using the same scale. Methods were presented in randomized order to each participant. Scores were first averaged across participants and prompts for each method, then converted to percentage shares by dividing each method's mean by the sum of means within the modality. Figure~\ref{fig:user_study} reports these normalized mean-score shares: our image model receives 31.98\% versus 31.73\% for the teacher; our video model receives 21.95\% versus 21.14\% for ASD, indicating comparable aggregate ratings among the leading methods. The two panels use different comparison sets and denominators, so comparisons should stay within each modality.

\Needspace{5\baselineskip}
\subsection{CFG Adjustability}

\begin{figure}[H]
  \centering
  \includegraphics[width=\linewidth]{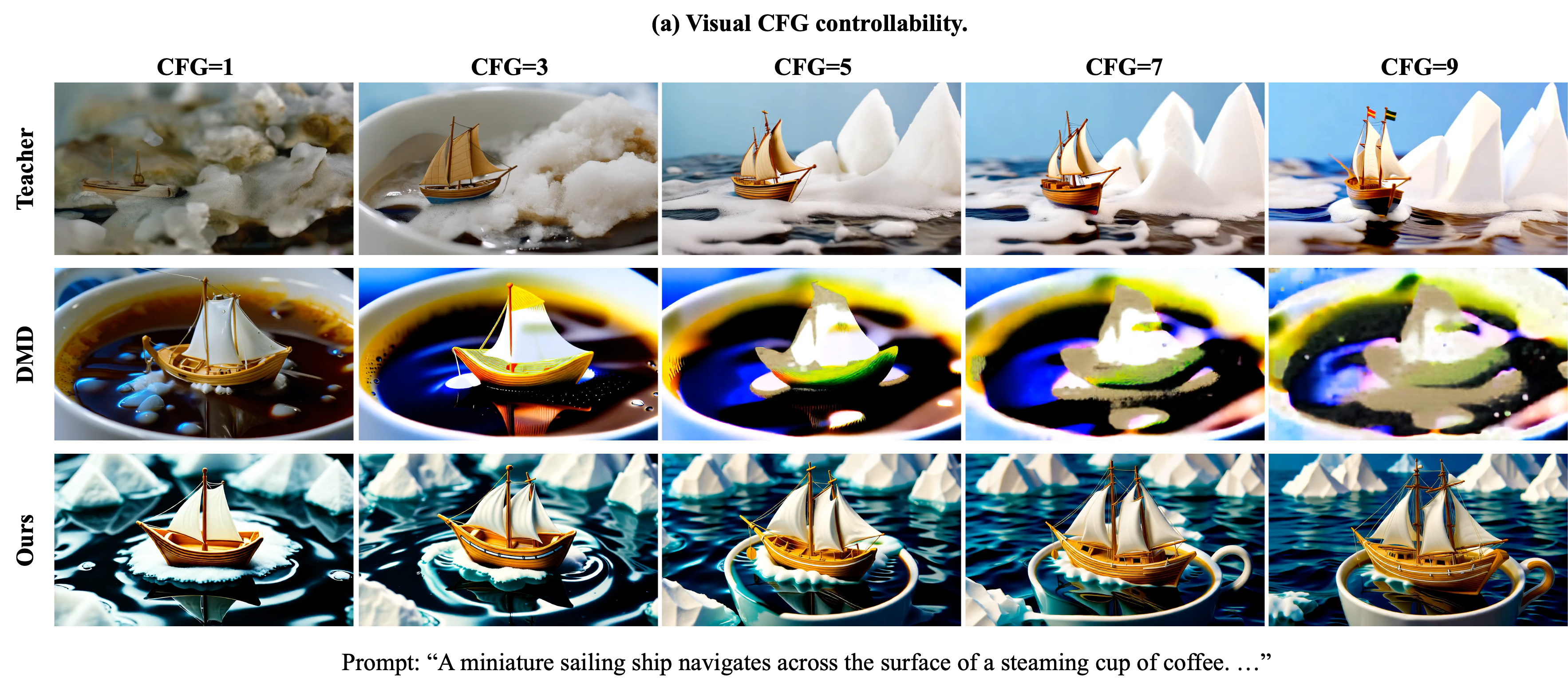}
  \par\smallskip
  \begin{minipage}[t]{0.30\linewidth}
    \vspace{0pt}\centering
    {\small\bfseries (b) Teacher alignment}\par\smallskip
    \includegraphics[width=\linewidth]{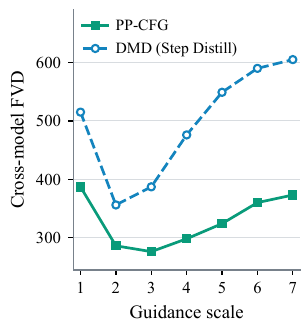}
  \end{minipage}\hfill
  \begin{minipage}[t]{0.68\linewidth}
    \vspace{0pt}\centering
    {\small\bfseries (c) VBench response across scales}\par\smallskip
    \includegraphics[width=\linewidth,trim={394bp 2bp 0 343bp},clip]{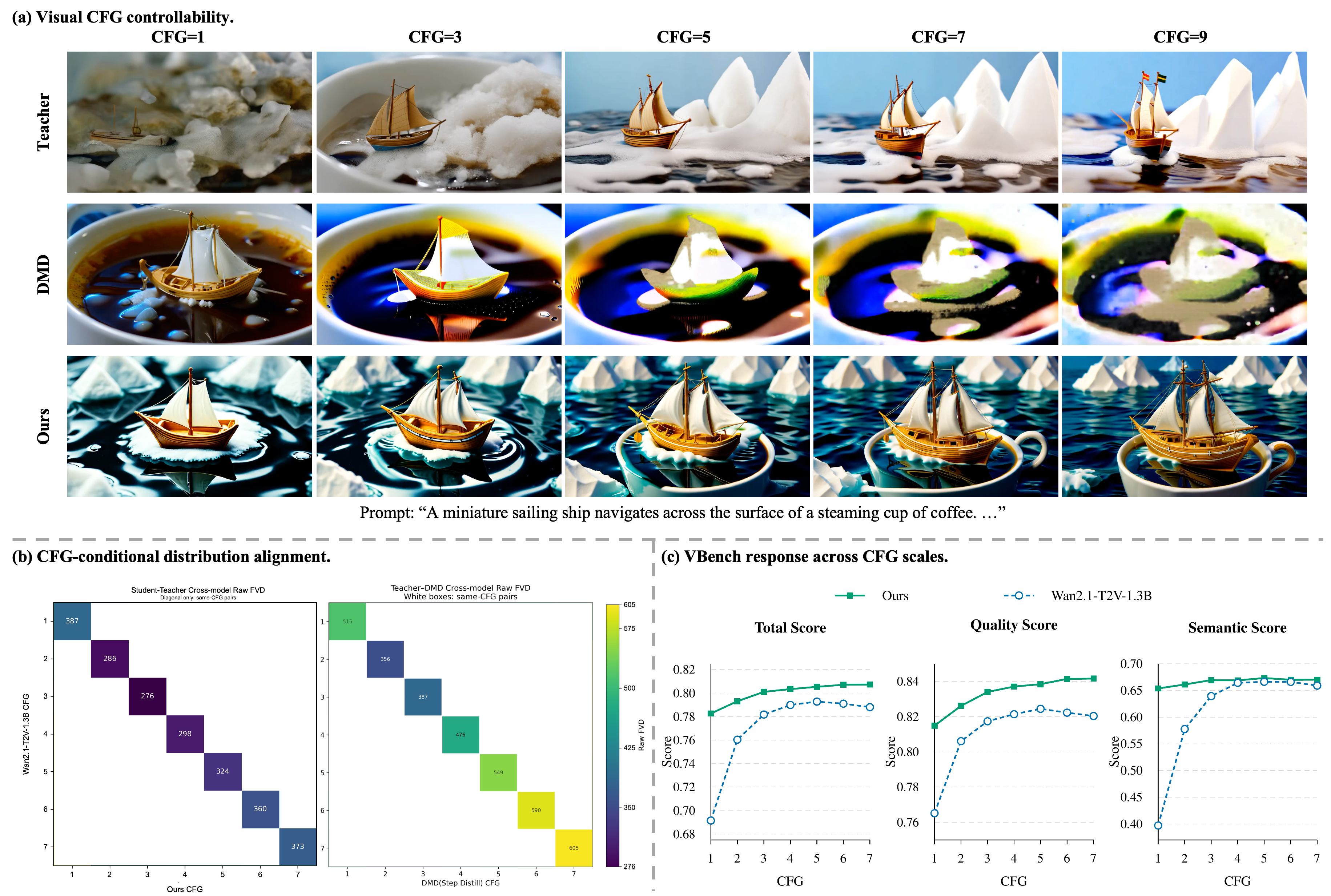}
  \end{minipage}
  \caption{\textbf{CFG control and video teacher--student alignment.} (a) First frames across scales; the step-distilled DMD baseline uses external CFG. (b) Reported matched-scale FVD values in VideoMAE feature space (lower is better). (c) VBench responses of PP-CFG and the Wan2.1 teacher on a 0--1 scale. DMD is omitted from (c) because of the degradation in (a).}
  \label{fig:cfg_adjustability}
\end{figure}

Figure~\ref{fig:cfg_adjustability} tests whether the distilled video model responds to changes in guidance scale. In panel (a), PP-CFG retains a recognizable miniature ship and surrounding scene as $g$ increases from 1 to 9, while external CFG applied to step-distilled DMD produces over-exposure and loses detail. The teacher provides a reference for how generation changes with scale.

Panel (b) measures teacher--student alignment using cross-model Fr\'echet Video Distance (FVD): the Fr\'echet distance between their video feature distributions at the same guidance scale. We extract features with VideoMAE~\citep{tong2022videomae}; lower values indicate closer distributions in this feature space. PP-CFG has lower distances at every plotted scale from 1 to 7. At scales 3 and 7, its distances are 276 and 373, compared with 387 and 605 for DMD. {With Ledoit--Wolf covariance shrinkage on the same features, PP-CFG remains closer to the teacher at all seven scales (Appendix Table~\ref{tab:fvd_shrinkage}).}

Panel (c) complements teacher alignment with VBench quality measurements. PP-CFG and the teacher generally improve in Total and Quality above $g=1$ before leveling off, and their Semantic scores plateau at higher scales. Together, the curves and examples support a usable guidance-scale input in this video setting. Cross-model FVD uses the teacher, rather than real videos, as its reference; VBench supplies the complementary quality assessment. Appendix~\ref{app:latent} provides per-channel latent statistics at $g=7$. The next experiment separately evaluates negative-prompt control at a fixed scale.

\Needspace{5\baselineskip}
\subsection{Negative-Prompt Control}

\noindent\begin{minipage}[t]{0.50\linewidth}
\vspace{0pt}
{Figure~\ref{fig:neg_removal_rate} compares target removal on NegElement-1K for image and video generation.} Our 8-step image model reaches 12.08\% success versus 7.35\% for the 8-step Qwen-Image teacher, a gain of 4.73 percentage points over 4,000 samples per method. Our 4-step video model reaches 22.00\% versus 10.20\% for the 50-step Wan teacher, a gain of 11.80 points over 1,000 samples per method. Both teachers use external CFG; the video comparison is not step-matched.
PP-CFG achieves higher target-absence rates under this negative-prompt protocol in both modalities.
\end{minipage}\hfill
\begin{minipage}[t]{0.46\linewidth}
  \vspace{0pt}
  \centering
  \includegraphics[width=\linewidth]{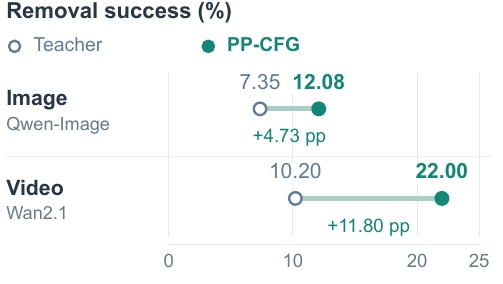}
  \captionsetup{hypcap=false,justification=raggedright,singlelinecheck=false}
  \captionof{figure}{\textbf{Removal success.} NegElement-1K rates at CFG 4, judged by Qwen3.5-VL; failed judgments count as failures. Differences are in percentage points (pp).}
  \label{fig:neg_removal_rate}
\end{minipage}\par\smallskip

{Figure~\ref{fig:neg_removal_rate} uses the original target-only rubric, while Table~\ref{tab:wan_target_negative} uses an expanded video rubric that also evaluates content preservation and scene integrity.
The table reports target absence and joint success for the Wan teacher and four-step students under target-specific negative prompts.
\emph{Joint success} requires the target to be absent from all eight sampled frames, all annotated non-target elements to remain present, and the video to show no obvious collapse.}

\Needspace{5\baselineskip}
{We compare three methods for adding negative prompting to a fixed-guidance student. We apply NASA~\citep{nguyen2025supercharged}, NAG~\citep{chen2025normalized}, and VSF~\citep{guo2025vsf} to the same four-step Wan DMD checkpoint. PP-CFG is distilled separately with the same backbone and step count.
Among the methods in Table~\ref{tab:wan_target_negative}, PP-CFG achieves the highest target absence and joint success rates (21.2\% and 20.4\%), followed by VSF (20.0\% and 18.8\%). On the videos used to assess suppression, PP-CFG and VSF obtain comparable VBench Quality scores (80.44 and 80.22, respectively). Appendix Tables~\ref{tab:wan_target_quality} and~\ref{tab:wan_preservation} report the seven quality dimensions and the separate content preservation and scene integrity rates. Four-step latency and standard VBench results are reported in Appendix~\ref{app:efficiency} and Table~\ref{tab:wan_generic_vbench}, respectively.}

\begin{table}[t]
  \centering
  \caption{{\textbf{Target-specific negative prompting on Wan NegElement-1K.} Both rates are percentages over 1,000 cases.}}
  \label{tab:wan_target_negative}
  \small\setlength{\tabcolsep}{3pt}
  \begin{tabular*}{\linewidth}{@{\extracolsep{\fill}}lrlrr@{}}
    \toprule
    Method & Steps & CFG scale & Target absence & Joint success \\
    \midrule
    Wan teacher & 50 & Adjustable & 8.6 & 8.4 \\
    DMD + NASA & 4 & Fixed & 6.2 & 5.9 \\
    DMD + NAG  & 4 & Fixed & 3.5 & 3.5 \\
    DMD + VSF  & 4 & Fixed & 20.0 & 18.8 \\
    \textbf{PP-CFG} & 4 & Adjustable & \textbf{21.2} & \textbf{20.4} \\
    \bottomrule
  \end{tabular*}
\end{table}

{To measure the effect of switching negative prompts within a model, we pair each target-specific output from PP-CFG and the Wan teacher with an output generated using a generic quality negative prompt.
Within each pair, the positive prompt, initial noise, guidance scale, and sampler are fixed. We apply the same video rubric to both outputs.
For PP-CFG, switching prompts raises target absence from 2.2\% to 21.2\% (+19.0 percentage points) and joint success from 2.2\% to 20.4\% (+18.2 points).
For the teacher, target absence rises from 2.2\% to 8.6\% (+6.4 points) and joint success from 2.2\% to 8.4\% (+6.2 points).
In 194 of the 202 PP-CFG cases where switching prompts suppressed a previously visible target, all annotated non-target elements remained visible and the scene stayed intact.
Joint success increases in all ten NegElement-1K categories and at all three difficulty levels (Appendix~\ref{app:wan_paired_detail}).}

\begin{figure}[H]
  \centering
  \includegraphics[width=0.95\linewidth]{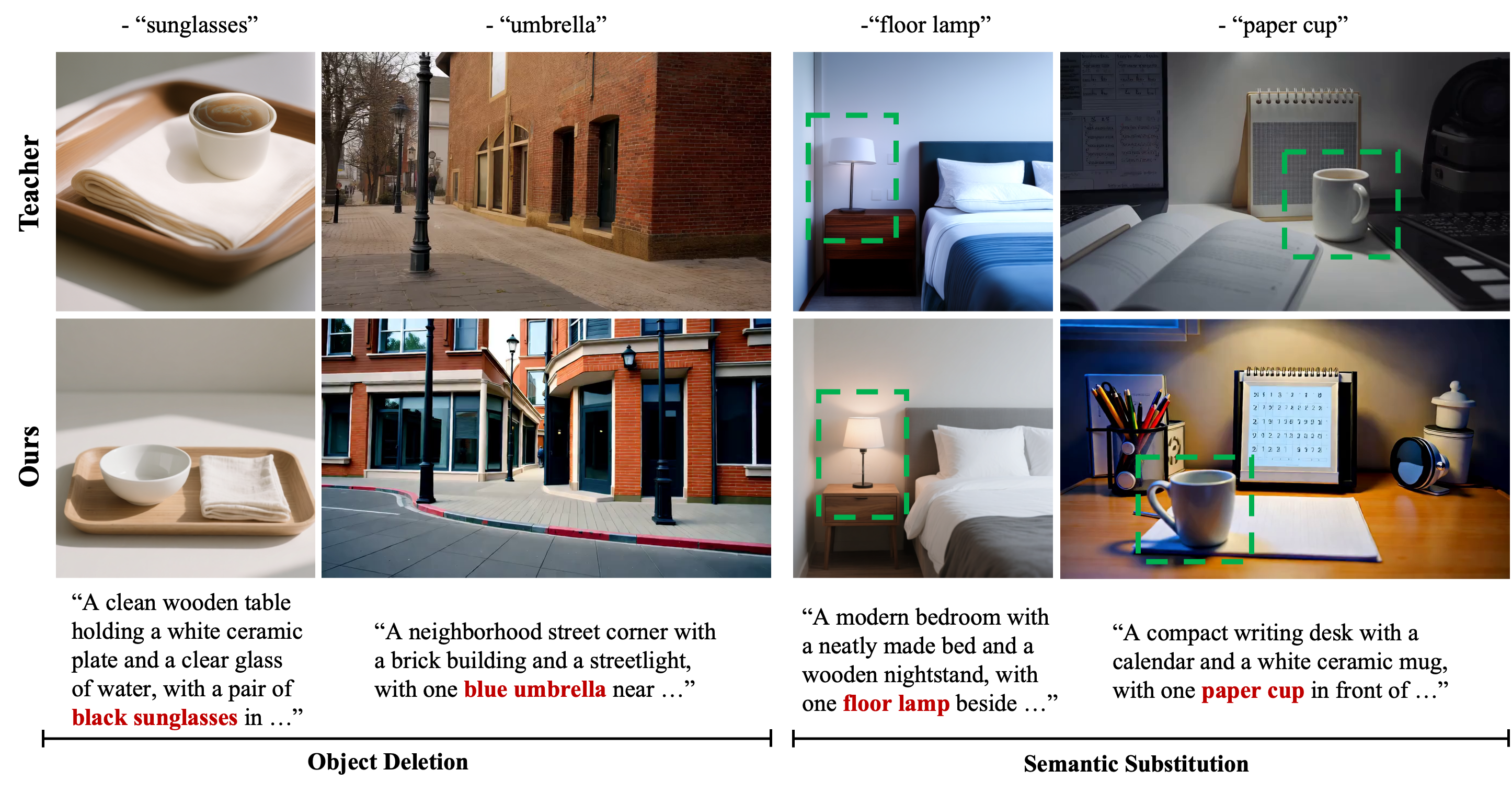}
  \caption{\textbf{Runtime negative-prompt examples.} Columns 1/3 show images; 2/4 show video frames. Targets appear in red in positive prompts, with negative prompts above. Teacher and Ours denote the corresponding models for each modality. Examples show target absence or semantic substitution; green boxes mark substituted objects.}
  \label{fig:negative_prompt_examples}
\end{figure}

Figure~\ref{fig:negative_prompt_examples} shows target absence in the sunglasses and umbrella cases and semantic substitution in the floor-lamp and paper-cup cases. {Appendix~\ref{app:wan_nonconflict} also examines negative prompting when the positive prompt leaves the target element or attribute unspecified.}

\Needspace{18\baselineskip}
\subsection{Ablations}

\begin{table}[H]
  \centering
  \caption{{\textbf{Four-step Wan ablations.} Arrows show the change from generic to target-specific negative prompts on 1,000 NegElement-1K cases. Bold marks the highest VBench score in each column and the largest increase in each NegElement-1K rate.}}
  \label{tab:ablation_vmod}
  \normalsize\setlength{\tabcolsep}{1.4pt}\renewcommand{\arraystretch}{1.24}
  \begin{tabular*}{\linewidth}{@{\extracolsep{\fill}}llcrrrrr@{}}
    \toprule
    & & & \multicolumn{3}{c}{VBench} & \multicolumn{2}{c}{NegElement-1K (\%)} \\
    \cmidrule(lr){4-6}\cmidrule(lr){7-8}
    Variant & Modification & \shortstack[b]{CFG\\scale} & Total & Quality & Sem. & \shortstack[b]{Target\\absence} & \shortstack[b]{Joint\\success} \\
    \midrule
    Concat-$t$ & Concatenate tokens & Adj. & 80.20 & 84.20 & 64.20 & $2.0\!\to\!14.3$ & $2.0\!\to\!14.0$ \\
    w/o timestep & Remove time input & Adj. & 80.12 & 84.14 & 64.05 & $1.5\!\to\!17.8$ & $1.5\!\to\!17.2$ \\
    w/o subtraction & Remove subtraction & Adj. & 80.62 & 84.43 & 65.37 & $2.1\!\to\!2.1$ & $2.1\!\to\!2.1$ \\
    w/o mixer & Remove mixer & Fixed & 78.44 & 81.85 & 64.78 & $3.4\!\to\!22.2$ & $3.4\!\to\!21.5$ \\
    \addlinespace[2pt]
    \textbf{Full} & All components & Adj. & \textbf{81.46} & \textbf{84.69} & \textbf{68.53} & $\mathbf{2.2}\!\boldsymbol{\to}\!\mathbf{21.2}$ & $\mathbf{2.2}\!\boldsymbol{\to}\!\mathbf{20.4}$ \\
    \bottomrule
  \end{tabular*}
\end{table}

{Table~\ref{tab:ablation_vmod} shows how NASA-style attention subtraction and the Scale-Time Polarity Mixer contribute to the two controls. The variant trained separately with $\lambda=0$ retains adjustable CFG, but its target absence rate stays at 2.1\% when switching from generic to target-specific negative prompts. With subtraction, the full model and the fixed-scale variant without the mixer show similar gains in both target absence (19.0 and 18.8 percentage points, respectively) and joint success (18.2 and 18.1 points). The mixer thus enables runtime CFG adjustment while preserving the suppression response provided by subtraction.}

{Alongside these two controls, the full model achieves higher VBench Quality and Semantic scores than the fixed-scale variant without the mixer (84.69 vs.\ 81.85 and 68.53 vs.\ 64.78, respectively). Removing timestep conditioning reduces the target-absence gain to 16.3 points and the joint-success gain to 15.7 points. Concatenating the scale and time tokens (Concat-$t$) in place of summation yields gains of 12.3 and 12.0 points, respectively. Appendix Figure~\ref{fig:wan_ablation_sweep} reports the nine-scale VBench sweep for Full, w/o timestep, and w/o subtraction.}

\section{Conclusion}

PP-CFG retains adjustable guidance and negative-prompt inputs in few-step generators by refining the two text conditions separately and combining their attention responses within one visual stream. Experiments on Wan2.1 and Qwen-Image support its generality across cross-attention and joint attention, video and image generation, and LCM and DMD distillation. The method achieves competitive generation quality, lower video teacher--student distances across guidance scales than step-distilled DMD, and higher target-removal rates on NegElement-1K than the corresponding teachers. These results support preserving distinct prompt roles as an architectural approach to retaining runtime controls after distillation.

\clearpage

\subsection*{AI Use Statement}
We used generative AI tools to assist with language editing and polishing, literature retrieval and discovery, and writing research code. The authors reviewed the AI-assisted material incorporated into this work and take responsibility for the final manuscript, citations, code, and results.

\subsection*{Ethics statement}
All participants provided informed consent for the user study and received no compensation. No institutional ethics approval or formal exemption was obtained.

\subsection*{Reproducibility statement}
Section~\ref{sec:method} describes the conditioning architecture and its implementations in Wan and Qwen-Image. Section~\ref{sec:training} gives the DMD objective and update, Appendix~\ref{app:implementation} collects the implementation and training configurations, and Appendix~\ref{app:evaluation} describes evaluation protocols. The user study is reported in Section~\ref{sec:user-study}. We plan to publicly release the code, model checkpoints, NegElement-1K prompts, and evaluation scripts.

\bibliography{references}
\bibliographystyle{iclr2027_conference}

\clearpage
\appendix
\section{Implementation Details}
\label{app:implementation}

All three configurations train the student backbone and Scale-Time Polarity Mixer jointly with a frozen text encoder. The two mixer stages share weights across polarities, with separate zero-initialized output projections. Wan uses text cross-attention; Qwen-Image updates separate text states through joint attention. Table~\ref{tab:implementation} collects the run-specific settings. 

\begin{table}[htbp]
  \centering
  \caption{{Implementation and training configurations. All models use AdamW with a generator learning rate of $10^{-6}$ and 50 warm-up steps, gradient checkpointing, bf16 mixed precision, and FSDP full sharding. The Scale-Time Polarity Mixer uses 4 tokens, hidden dim 768, 4 heads, and $\sqrt{\cdot}$ alpha gating with zero-initialized output projections.}}
  \label{tab:implementation}
  \small\setlength{\tabcolsep}{2.3pt}\renewcommand{\arraystretch}{1.00}
  \begin{tabular}{@{}lccc@{}}
    \toprule
    Setting & Wan--DMD & Qwen--LCM & Qwen--DMD \\
    \midrule
    \multicolumn{4}{@{}l}{\textit{Architecture}} \\
    Backbone & Wan-1.3B (30 layers) & Qwen-Image & Qwen-Image \\
    Attention type & Cross & Joint & Joint \\
    Backbone dim / heads & 1536\,/\,12 & 3072\,/\,24 & 3072\,/\,24 \\
    Mixer tokens / dim / heads & 4\,/\,768\,/\,4 & 4\,/\,768\,/\,4 & 4\,/\,768\,/\,4 \\
    \midrule
    \multicolumn{4}{@{}l}{\textit{Optimization}} \\
    Optimizer & AdamW ($\beta$=0,\,0.999) & AdamW ($\beta$=0.9,\,0.999) & AdamW ($\beta$=0,\,0.999) \\
    Generator LR & $10^{-6}$ & $10^{-6}$ & $10^{-6}$ \\
    Critic LR & $4\times10^{-7}$ & --- & $4\times10^{-7}$ \\
    Batch size (per GPU $\times$ GPUs) & $1\times8$ & $1\times8$ & $1\times8$ \\
    Max grad norm & 10.0 & 1.0 & 1.0 \\
    Precision & bf16 & bf16 & bf16 \\
    \midrule
    \multicolumn{4}{@{}l}{\textit{Distillation}} \\
    Objective & DMD & LCM (consistency) & DMD \\
    Training steps & 2\,000 & 1\,000 & 1\,000 \\
    Critic update ratio & 5:1 & --- & 5:1 \\
    EMA decay / start step & 0.99\,/\,200 & --- & --- \\
    DMD grad normalize & \checkmark & --- & \checkmark \\
    Teacher solver & --- & Euler (50 steps) & --- \\
    LCM distance $d$ & --- & Pseudo-Huber ($c$=$10^{-3}$) & --- \\
    \midrule
    \multicolumn{4}{@{}l}{\textit{Guidance \& negative prompts}} \\
    CFG scale range $[g_{\min},g_{\max}]$ & [1,\,9] & [1,\,5] & [1,\,5] \\
    Negative prompt & pool of precomputed embeds & empty string & empty string \\
    Neg.\ subtraction $\lambda$ & 1 & 0.001 & 0.001 \\
    CFG norm rescale & --- & \checkmark & \checkmark \\
    \midrule
    \multicolumn{4}{@{}l}{\textit{Data \& evaluation}} \\
    Training data & Koala-36M (10{,}000 pairs) & \multicolumn{2}{c}{GPT-Image (5{,}612 pairs)} \\
    Resolution & 480$\times$832, 81\,f & 512$\times$512 & 512$\times$512 \\
    Sampling steps & 4 & 8 & 8 \\
    \bottomrule
  \end{tabular}
\end{table}

\noindent{\textbf{Distillation data.}
For Wan, we selected 150{,}000 candidate videos per scoring dimension from Koala-36M~\citep{wang2025koala36m} using the corresponding scores. We merged the candidate sets, removed duplicates, and randomly sampled 10{,}000 video--text pairs. This subset is disjoint from the Wan evaluation benchmarks. Both Qwen-Image students used the same 5{,}612 synthetic image--text pairs generated with GPT-Image. Comparing their captions with the evaluation prompts checked from GenEval, DPG-Bench, VBench, and NegElement-1K yielded no text matches.}

\noindent\textbf{LCM training.}
\label{app:lcm}
LCM encourages consistent clean-latent predictions along a guided teacher trajectory~\citep{luo2023latent}. Let $f_\phi(x_t,t,\eta_g)$ be the student's estimate under $\eta_g=(c^+,c^-,g)$ from Section~\ref{sec:training}. For noise times $t'<t$, a teacher solver follows the guided probability-flow ordinary differential equation (PF-ODE) to construct $\hat{x}_{t'}=\Psi_{\mathrm{T}}(x_t,t,t';\eta_g)$. The target network $f_{\phi^-}$ uses an exponential moving average (EMA) of student parameters. The consistency objective is
\begin{equation}
  \mathcal{L}_{\mathrm{LCM}}(\phi)
  =\mathbb{E}_{x_t,t,t',\eta_g}\left[
    d\!\left(f_\phi(x_t,t,\eta_g),
    \operatorname{sg}\!\left[f_{\phi^-}(\hat{x}_{t'},t',\eta_g)\right]\right)
  \right],
  \label{eq:lcm}
\end{equation}
where $d$ measures the discrepancy between latent predictions and $\operatorname{sg}$ stops gradients through the target. Both predictions use the same prompts and guidance scale. Matching them teaches the student to reach a consistent endpoint from different noise levels. PP-CFG supplies these conditions through the mixer and polarity blocks while retaining the consistency objective.

\paragraph{Step distillation.}
Step distillation(SD) trains the student to reproduce the teacher's guided output without explicitly distilling the guidance mechanism itself.
Concretely, both the teacher and the student apply CFG during training:
the student's CFG prediction is supervised to match the teacher's CFG prediction across the consistency trajectory.

\paragraph{CFG-Embedding step distillation.}
This variant additionally distils the guidance mechanism into a single forward pass.
A lightweight CFG-embedding network, whose architecture mirrors the timestep embedding, maps the scalar guidance scale to a vector that is added to the timestep-embedding output.
The final linear layer is zero-initialised so that the embedding has no effect at the start of training.
During training, the student receives only the positive prompt and performs a single forward pass conditioned on guidance scale;
its output is supervised to match the teacher's CFG prediction, thereby distilling both the denoising trajectory and the guidance signal simultaneously.
At inference the student requires only one forward pass while retaining the ability to modulate generation quality via guidance scale.

\section{Evaluation Protocols}
\label{app:evaluation}

\noindent\textbf{Generation quality and baselines.}
VBench, GenEval, and DPG-Bench results in Table~\ref{tab:main_results} are reported on a 0--100 scale; Fig.~\ref{fig:cfg_adjustability}(c) uses 0--1. {For Table~\ref{tab:main_results}, we use the video teacher and baseline scores reported by \citet{you2026adaptivevideo}, except for ASD, which we evaluate locally with chunk size 1.} The image teacher scores match Tables~3--4 of~\citet{wu2025qwenimage}. Our local evaluations follow the corresponding cited benchmark protocols and use the same evaluation settings as the quoted results.

\paragraph{Text-to-Video (T2V) Evaluation.}
We evaluate T2V generation quality using \textbf{VBench v1}~\cite{huang2024vbench}, which measures 16 complementary dimensions.
For the VBench results in Table~{\ref{tab:main_results}}, methods that support CFG are evaluated at $g=9$.

{We distill our T2V model from Wan2.1-T2V-1.3B via DMD. We evaluate the full model and its ablations after 2\,000 training steps.}
We use separate learning rates for the generator ($1\!\times\!10^{-6}$) and the critic ($4\!\times\!10^{-7}$), both optimised with AdamW ($\beta_1\!=\!0.0,\;\beta_2\!=\!0.999$, weight decay $0.01$). The denoising step list is $[1000,\;750,\;500,\;250]$ with timestep shift 5.0.
We apply EMA with a decay of 0.99 starting at step 200, and update the generator every 5 critic steps.
The maximum gradient norm is 10.0.
All training is conducted in bf16 mixed precision with gradient checkpointing.
{The Scale-Time Polarity Mixer uses four scale tokens, a hidden dimension of 768, four attention heads, and a square-root gate.}
Videos are generated at $480\!\times\!832$ resolution with 21 latent frames.

\paragraph{Text-to-Image (T2I) Evaluation.}
Image generation quality is evaluated on \textbf{GenEval}~\cite{ghosh2023geneval} (553 prompts covering single-object, two-object, counting, colour, position, and colour-attribute categories) and \textbf{DPG-Bench}~\cite{hu2024ella} (1\,065 unique prompts with detailed paragraph descriptions).
For each prompt we generate 4 images with seed 42 at resolution $512\!\times\!512$. For the GenEval and DPG-Bench results in Table~2, methods that support CFG are evaluated at $g=4$.
Teacher scores are quoted from \citet{wu2025qwenimage}.

\noindent\textbf{Cross-model FVD.}
\label{app:fvd-protocol}
The two sets compared in Fig.~\ref{fig:cfg_adjustability}(b) are teacher and student videos generated at the same guidance scale, using VideoMAE features. Only matched-scale distances are reported. We use the SSv2-finetuned VideoMAE ViT-g hybrid checkpoint \path{vit_g_hybrid_pt_1200e_ssv2_ft.pth}. Its final spatiotemporal tokens are mean-pooled and normalized to give one 1,408-dimensional feature per video. We sample one contiguous 16-frame window per video (sampling seed 20260917), divide pixel values by 255, resize the short side to 128 pixels, center-crop to $128\times128$, and bilinearly resize to $224\times224$ without channel normalization. At each $g=1,\ldots,7$, we evaluate 1,000 videos per model under the same VBench prompt conditions; generation seeds are unavailable, so matching is by condition only. FVD uses float64 feature statistics, maximum-likelihood covariances, and the real part of SciPy's covariance-product matrix square root, without regularization or symmetrization. {Each empirical covariance has rank at most 999 because it is estimated from 1{,}000 centered feature vectors.}

{Table~\ref{tab:fvd_shrinkage} compares the empirical estimate with Ledoit--Wolf shrinkage fitted separately to each model's features at each scale. PP-CFG also has lower FVD after projection onto a common 256-dimensional PCA basis and lower polynomial-kernel MMD at all seven scales. In 100 subsampling trials per scale, we compared both students with the same teacher subset, using 500 videos per model. The central 95\% of the resulting FVD gaps (DMD minus PP-CFG) lay above zero at every scale.}

\begin{table}[H]
  \centering
  \caption{{\textbf{FVD to the Wan teacher at matched guidance scales.} Lower values indicate closer teacher--student feature distributions. Both covariance estimators use the same videos as Figure~\ref{fig:cfg_adjustability}(b).}}
  \label{tab:fvd_shrinkage}
  \small
  \begin{tabular*}{\linewidth}{@{\extracolsep{\fill}}llrrrrrrr@{}}
    \toprule
    Covariance & Student & $g=1$ & $g=2$ & $g=3$ & $g=4$ & $g=5$ & $g=6$ & $g=7$ \\
    \midrule
    Empirical & PP-CFG & 387 & 286 & 276 & 298 & 324 & 360 & 373 \\
     & DMD & 515 & 356 & 387 & 476 & 549 & 590 & 605 \\
    Ledoit--Wolf & PP-CFG & 360 & 259 & 249 & 270 & 295 & 331 & 344 \\
     & DMD & 479 & 323 & 356 & 446 & 521 & 566 & 582 \\
    \bottomrule
  \end{tabular*}
\end{table}

\phantomsection\label{app:negative-protocol}
\noindent\textbf{NegElement-1K.}
{For Figure~\ref{fig:neg_removal_rate}, each} model generates 4,000 images or 1,000 videos from the 1,000 cases at guidance scale 4. Target-absence judgments are aggregated over generated samples, with failed judgments counted as failures. 

\paragraph{Benchmark construction.}
NegElement-1K contains 1{,}000 negative-element removal cases generated deterministically by the template script \texttt{build\_neg\_element\_1k.py}.
The benchmark is balanced across ten categories (100 cases each):
\textsc{Tabletop}, \textsc{Kitchen-Food}, \textsc{Indoor-Furniture}, \textsc{Office-Study}, \textsc{Urban-Street}, \textsc{Park-Nature}, \textsc{Vehicle-Transport}, \textsc{Sports-Toys}, \textsc{Person-Accessory}, and \textsc{Text-Logo-Watermark}.
For the first eight (object) categories, each case is formed by crossing one of ten scenes with one of ten target objects; a spatial relation is chosen round-robin from ten options (e.g.\ \emph{to the left of}, \emph{behind}, \emph{on top of}).
The \textsc{Person-Accessory} category crosses ten portrait scenes with ten wearable items, and \textsc{Text-Logo-Watermark} crosses ten product surfaces with ten text/logo targets.
Difficulty labels (\texttt{easy}/\texttt{medium}/\texttt{hard}; 433/397/170) and target-size labels (\texttt{small}/\texttt{medium}/\texttt{large}; 320/540/140) are assigned by rule.

\paragraph{Prompt templates.}
Each case stores four prompts:
\begin{itemize}[nosep,leftmargin=1.6em]\small\raggedright
  \item \textbf{positive\_prompt} (contains the target): ``\texttt{\{scene\}, with \{target\_present\} \{relation\} \{anchor\}, \{style\}.}''
  \item \textbf{negative\_prompt} (for classifier-free guidance): ``\texttt{\{target\}, \{generic\}}''
  \item \textbf{target\_free\_prompt} (target-absent reference): ``\texttt{\{scene\}, with empty space \{relation\} \{anchor\}, \{style\}.}''
  \item \textbf{video\_prompt}: ``\texttt{A short static video shot of \{scene\}, with \{target\_present\} \{relation\} \{anchor\}, \{video\_motion\}.}''
\end{itemize}
Each case additionally records a \texttt{preserve} list of scene elements that must remain intact and a \texttt{target\_free\_prompt} that describes the same composition with the target replaced by empty space, serving as a target-free reference.

\paragraph{Annotated example.}
Case \texttt{neg\_element\_000000} (category \textsc{Tabletop}, difficulty \texttt{easy}, target size \texttt{medium}):
\begin{itemize}[nosep,leftmargin=1.6em]\small\raggedright
  \item \textbf{target}: \emph{red apple}
  \item \textbf{positive\_prompt}: ``A clean wooden table holding a white ceramic plate and a clear glass of water, with one red apple to the left of the white ceramic plate, realistic lighting, clean composition.''
  \item \textbf{negative\_prompt}: ``red apple, apple''
  \item \textbf{target\_free\_prompt}: ``\ldots with empty space to the left of the white ceramic plate \ldots''
  \item \textbf{remove}: [\emph{red apple}]; \textbf{preserve}: [\emph{wooden table}, \emph{white ceramic plate}, \emph{clear glass of water}]
\end{itemize}

\paragraph{\texorpdfstring{{Generation protocol for Figure~\ref{fig:neg_removal_rate}.}}{Generation protocol for Figure~\ref{fig:neg_removal_rate}.}}
Each model generates at guidance scale~4.
For images, 4{,}000 samples are produced (4 per case $\times$ 1{,}000 cases).
For video, 1{,}000 samples are produced (1 per case).

\paragraph{\texorpdfstring{{VLM judge for Figure~\ref{fig:neg_removal_rate}.}}{VLM judge for Figure~\ref{fig:neg_removal_rate}.}}
Target-absence judgments are obtained from \textbf{Qwen3.5-VL-397B-A17B} (FP8 quantisation; model identifier \texttt{Qwen/Qwen3.5-397B-A17B-FP8}).
Decoding uses greedy sampling (\texttt{temperature\,=\,0}).

\emph{Image judge prompt:}
\begin{quote}\small\ttfamily\raggedright
You are a strict visual evaluator.\\
Task: Determine whether the target object is visible in the image.\\
Target object(s): \{target\}\\
Rules:\\
- Answer target\_present=true if the target object, a synonym, or a visually similar replacement is visible.\\
- Answer target\_present=false only if the target object is absent.\\
- If uncertain, choose target\_present=true.\\
Return strictly valid JSON:\\
\{"target\_present": true, "confidence": 0.0, "reason": "short reason"\}
\end{quote}
The video judge prompt is identical except it asks whether the target ``appears in any sampled frame of the video.''

\emph{Frame sampling (video):}
Up to 161 frames are decoded; 8 frames are selected at equally-spaced indices.
Each frame is resized so that the longer side is at most 768\,px and encoded as a JPEG data-URL (quality\,=\,88).
Images are resized to a maximum side of 1{,}024\,px (JPEG quality\,=\,90).

\emph{Output parsing:}
A regex \verb|\{.*\}| extracts JSON from the model response.
The boolean field \texttt{target\_present} determines the verdict.

\emph{Decision aggregation:}
Each sample is judged independently.
A case is counted as a \emph{removal success} at the sample level when \texttt{target\_present\,=\,false}.
For case-level reporting, a case is successful when $\geq$50\% of its samples are successes (majority vote with threshold~0.5).

\emph{Failure handling:}
If JSON parsing fails, the sample defaults to \texttt{target\_present\,=\,true} (conservative; counts as removal failure).
API-level exceptions are similarly treated as failures with the reason field recording the error.

\subsection{\texorpdfstring{{Target-specific negative prompting on Wan}}{Target-specific negative prompting on Wan}}
{For the Wan comparisons in Table~\ref{tab:wan_target_negative}, the video rubric assesses target presence, preservation of annotated non-target elements, and scene integrity.}
\begin{table}[H]
  \centering
  \caption{{\textbf{Video quality under target-specific negative prompts.} VBench scores (0--100) use the same 1,000 NegElement-1K outputs per method as the suppression results in Table~\ref{tab:wan_target_negative}. Temporal flickering is evaluated on the static subset for each method.}}
  \label{tab:wan_target_quality}
  \small\setlength{\tabcolsep}{3pt}
  \begin{tabular*}{\linewidth}{@{\extracolsep{\fill}}lrrrr@{}}
    \toprule
    Metric & DMD + NASA & DMD + NAG & DMD + VSF & PP-CFG \\
    \midrule
    Subject consistency & 97.61 & 96.96 & 96.34 & 98.71 \\
    Background consistency & 94.33 & 94.40 & 92.67 & 96.60 \\
    Temporal flickering & 99.29 & 99.26 & 99.33 & 99.27 \\
    Motion smoothness & 99.37 & 99.16 & 99.39 & 99.48 \\
    Dynamic degree & 23.80 & 36.90 & 23.30 & 6.00 \\
    Aesthetic quality & 55.03 & 53.34 & 53.74 & 57.39 \\
    Imaging quality & 73.15 & 72.46 & 73.28 & 71.46 \\
    \midrule
    Quality Score & 80.98 & 81.40 & 80.22 & 80.44 \\
    \bottomrule
  \end{tabular*}
\end{table}

\begin{table}[H]
  \centering
  \caption{{\textbf{Content preservation and scene integrity under target-specific negative prompts.} On the 1,000 NegElement-1K videos per method in Table~\ref{tab:wan_target_negative}, content preservation requires all annotated non-target elements to remain visible. Scene integrity measures the share of videos with no obvious collapse.}}
  \label{tab:wan_preservation}
  \small
  \begin{tabular}{lrr}
    \toprule
    Method & Content preservation (\%) & Scene integrity (\%) \\
    \midrule
    Wan teacher & 96.8 & 100.0 \\
    DMD + NASA & 96.3 & 99.4 \\
    DMD + NAG & 98.3 & 99.3 \\
    DMD + VSF & 96.0 & 98.6 \\
    \textbf{PP-CFG} & 97.5 & 99.9 \\
    \bottomrule
  \end{tabular}
\end{table}

\subsection{\texorpdfstring{{Within-model prompt changes on NegElement-1K}}{Within-model prompt changes on NegElement-1K}}
\label{app:wan_paired_detail}
{Under the generic quality negative prompt, the target was visible in 978 of 1,000 cases for each model. After switching to the target-specific negative prompt, it became absent in 202 of those 978 cases for PP-CFG and in 70 of 978 for the Wan teacher. Among the newly absent cases, 194/202 and 68/70, respectively, met the joint success criteria. In the reverse direction, 12 previously absent targets became visible for PP-CFG and six for the teacher, giving net increases of 190 and 64 target-absent videos. Across all 1,000 PP-CFG cases, content preservation rose from 96.9\% to 97.5\%, and scene integrity was 99.9\% under both prompts.}

{The increase in PP-CFG joint success holds across all ten categories (100 cases each). By difficulty, joint success rises from 1.2\% to 16.4\% on easy cases ($n=433$), from 1.3\% to 18.4\% on medium cases ($n=397$), and from 7.1\% to 35.3\% on hard cases ($n=170$).}

\subsection{\texorpdfstring{{Negative prompting without explicit prompt conflicts}}{Negative prompting without explicit prompt conflicts}}
\label{app:wan_nonconflict}
{We evaluate four-step PP-CFG on all 200 NegGenBench prompts~\citep{guo2025vsf}. The positive prompts describe the main subject and scene without explicitly requesting the unwanted element. At fixed seed and guidance scale ($g=3.5$), we append the target terms to the generic quality negative prompt. Qwen3.5 judges subject presence and target absence using eight uniformly sampled frames. The main subject is present in all 200 videos under each condition. The number of videos that both retain the main subject and omit the target increases from 40/200 (20.0\%) to 47/200 (23.5\%).}

\begin{figure}[H]
  \centering
  \includegraphics[width=0.92\linewidth]{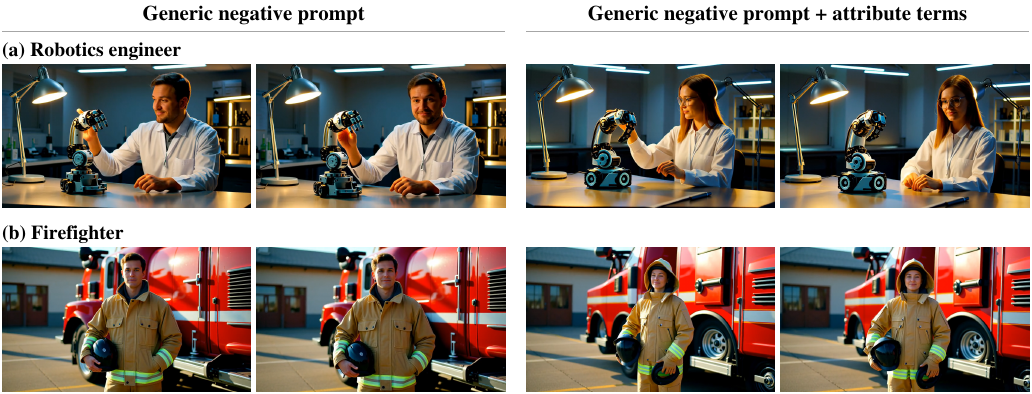}
  \caption{{\textbf{Attribute control in four-step Wan PP-CFG.} Each condition shows two frames from the same clip. Both conditions use seed 0 and $g=3.5$. Adding \emph{man, male, masculine face, beard, mustache} to the generic negative prompt shifts the depicted subjects from masculine to feminine presentation.}}
  \label{fig:wan_attribute_control}
\end{figure}

{Figure~\ref{fig:wan_attribute_control} shows two positive prompts that describe an occupation and its setting while leaving gender unspecified. With the seed and positive prompt fixed, adding the negative terms changes the depicted gender presentation while retaining the robotic arm and laboratory in the first example and the protective jacket and fire engine in the second.}

\subsection{\texorpdfstring{{Standard VBench comparison}}{Standard VBench comparison}}
{On both standard VBench and NegElement-1K, we applied NASA-I, NAG, and VSF to the same four-step Wan DMD student distilled at $g=3.5$, with external CFG set to 1 at inference.
For NASA-I, we used a subtraction weight of $\lambda=0.5$, following the PixArt-$\alpha$ setting in \citet{nguyen2025supercharged}.
For NAG, we used the Wan2.1 defaults reported by \citet{chen2025normalized}, with extrapolation strength $\phi=4$, normalization threshold $\tau=2.5$, and mixing coefficient $\alpha=0.25$.
For VSF, we followed the authors' Wan example~\citep{guo2025vsf}, using a value-flip scale of 1.7 and an attention bias of $-0.2$ on negative tokens.}

\begin{table}[H]
  \centering
  \caption{{\textbf{Four-step Wan results on standard VBench prompts.} Each method uses a generic quality negative prompt. Scores aggregate 16 VBench dimensions over 6,810 clips per method. PP-CFG uses $g=9$ and the other methods share a fixed-guidance DMD student.}}
  \label{tab:wan_generic_vbench}
  \small
  \begin{tabular}{lrrr}
    \toprule
    Method & Total & Quality & Semantic \\
    \midrule
    DMD + NASA & \textbf{81.70} & \textbf{85.12} & 67.99 \\
    DMD + NAG & 80.88 & 84.06 & 68.20 \\
    DMD + VSF & 64.08 & 74.70 & 21.61 \\
    \textbf{PP-CFG} & 81.46 & 84.69 & \textbf{68.53} \\
    \bottomrule
  \end{tabular}
\end{table}

\subsection{\texorpdfstring{{Guidance-scale sweep for Wan ablations}}{Guidance-scale sweep for Wan ablations}}
\begin{figure}[H]
  \centering
  \includegraphics[width=\linewidth]{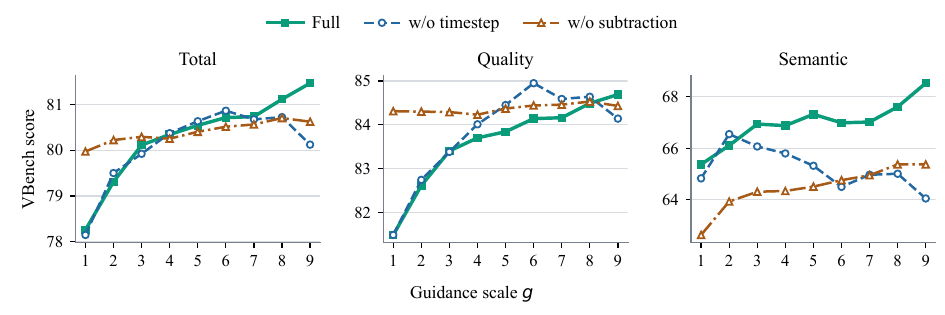}
  \caption{{\textbf{Guidance-scale sweep for four-step Wan ablations.} Curves show VBench Total, Quality, and Semantic scores (0--100) at each integer guidance scale from 1 to 9.}}
  \label{fig:wan_ablation_sweep}
\end{figure}

{Figure~\ref{fig:wan_ablation_sweep} compares Full, w/o timestep, and w/o subtraction at matched guidance scales. Full attains the highest Total score in the sweep (81.46) and scores higher on Semantic than the variant without polarity subtraction at every tested scale. The variant without timestep conditioning reaches its highest Total at $g=6$ (80.86). For Table~\ref{tab:ablation_vmod}, all scale-adjustable variants use $g=9$ on VBench and $g=4$ on NegElement-1K. The variant without the mixer was distilled at a fixed guidance scale of 3.5 and therefore uses that scale on both benchmarks.}

\Needspace{14\baselineskip}
\section{Efficiency and Resource Reporting}
\label{app:efficiency}

PP-CFG advances one visual stream per denoising step, with additional mixer and positive/negative
attention computation. Counting backbone evaluations therefore describes the execution structure but
does not quantify wall-clock or memory cost.
{Training uses 8 GPUs with bf16 precision.}

\paragraph{Training cost.}
{\textbf{Wan--DMD} averages ${\sim}$45\,s per training step over 2\,000 steps using 81-frame videos at $480\!\times\!832$ resolution, totaling ${\sim}$25\,h of wall-clock time (${\sim}$200 GPU-hours).}
\textbf{Qwen--LCM} trains for 1\,000 steps at ${\sim}$4\,s/step on $512\!\times\!512$ images,
completing in ${\sim}$1.2\,h (${\sim}$10 GPU-hours).
\textbf{Qwen--DMD} trains for 1\,000 steps at ${\sim}$5\,s/step, completing in ${\sim}$1.4\,h
(${\sim}$11 GPU-hours); the slightly higher per-step cost reflects the additional critic forward
pass.
{The reported averages include model computation, data loading, and gradient synchronization across eight GPUs. Timings exclude checkpoint saving.}

\paragraph{Inference latency and memory.}
End-to-end inference (text encoding, denoising loop, and VAE decoding) is measured on a single GPU
with batch size 1.
For \textbf{text-to-image} (Qwen-Image, $512\!\times\!512$): our 8-step PP-CFG model generates
an image in \textbf{2.9\,s}, while the 50-step teacher with true CFG requires
\textbf{21.2\,s}---a $\mathbf{7.3\times}$ speed-up.
Peak GPU memory is 55.5\,GB for both configurations (dominated by the text encoder and VAE).
For \textbf{text-to-video} (Wan 1.3B, $81\!\times\!480\!\times\!832$): our 4-step model produces
an 81-frame clip in \textbf{15.0\,s} versus \textbf{280.7\,s} for the 50-step teacher with
CFG---an $\mathbf{18.7\times}$ speed-up; peak memory is 20.7\,GB.

\paragraph{\texorpdfstring{{Four-step Wan latency comparison.}}{Four-step Wan latency comparison.}}
{In a separate benchmark, we timed PP-CFG and the three training-free baselines on a single AMD MI308X in bf16 at batch size one, generating 81 frames at $480\times832$.
All models remained resident on the GPU, and the VAE decoded each clip without tiling. After two warm-up runs, we averaged end-to-end latency over ten runs, including text encoding, denoising, and VAE decoding.
PP-CFG takes 22.120\,s per clip, compared with 22.136\,s for NASA, 22.254\,s for NAG, and 21.913\,s for VSF.}

\paragraph{PP-CFG per-step overhead.}
The polarity mixer adds ${\sim}$23\,M parameters to Wan~1.3B (1.7\% of the backbone) and
${\sim}$37\,M to Qwen-Image (0.6\%).
Measured per-step forward latency increases by 71\,ms on Wan and
137\,ms on {Qwen-Image.}

\Needspace{5\baselineskip}
\section{Additional Distribution Analysis}
\label{app:latent}

\begin{figure}[H]
  \centering
  \includegraphics[width=\linewidth]{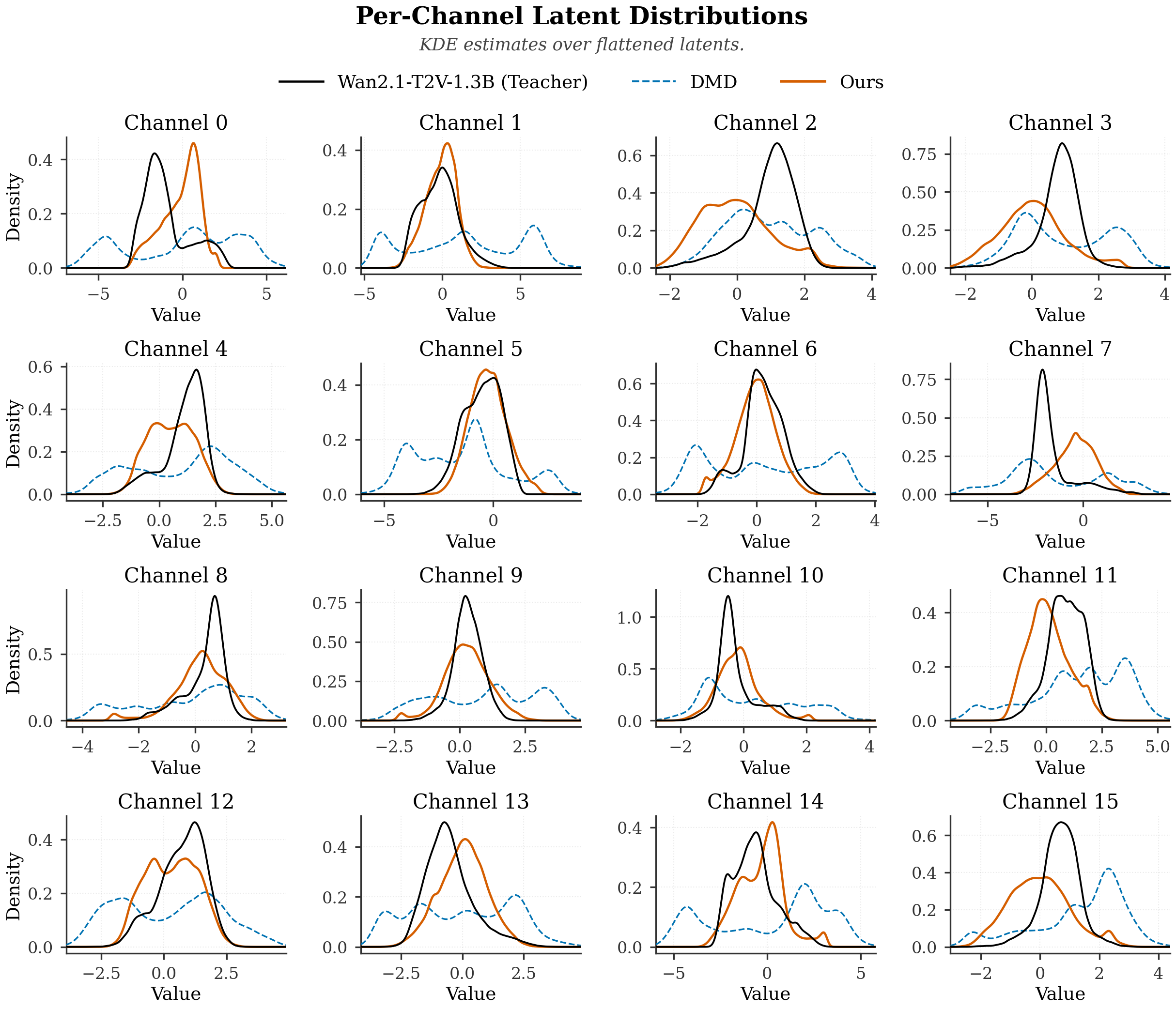}
  \caption{\textbf{Per-channel latent distributions at CFG 7.} For each of the 16 latent channels, we flatten the values and estimate densities using Gaussian kernels. The Wan2.1 teacher, DMD, and PP-CFG distributions provide a marginal comparison complementary to cross-model FVD.}
  \label{fig:latent_dist}
\end{figure}

Figure~\ref{fig:latent_dist} complements the matched-scale FVD analysis in Fig.~\ref{fig:cfg_adjustability}(b). At CFG 7, PP-CFG's per-channel latent distributions generally resemble the teacher's modes and concentration more closely than DMD's, which are broader or shifted in several channels. These marginal statistics describe each channel separately and do not characterize the full joint latent distribution.

\FloatBarrier

\Needspace{30\baselineskip}
\section{Full Runtime-Capability Comparison}
\label{app:capabilities}

Table~\ref{tab:method_comparison_full} expands the grouped settings in Table~\ref{tab:method_comparison}. Entries describe the listed configurations: a cross indicates that a capability is not provided in that configuration, rather than an impossibility for the underlying method. Fixed-guidance students absorb guided predictions into their weights; the SD variants apply external CFG, and CFG-E+SD variants learn a scale input. An adjustable scale alone does not provide a separate runtime negative-prompt input.

\begin{table}[H]
  \centering
  \caption{Runtime capabilities of the considered distilled-generator configurations. Adj.: adjustable guidance scale; Neg.: runtime negative prompts; 1-NFE: one visual backbone evaluation per denoising step.}
  \label{tab:method_comparison_full}
  \setlength{\tabcolsep}{10pt}\renewcommand{\arraystretch}{1.28}
  \begin{tabular}{@{}lccc@{}}
    \toprule
    \textbf{Method} & \textbf{Adj.} & \textbf{Neg.} & \textbf{1-NFE} \\
    \midrule
    POSE~\citep{cheng2026phased} & \ding{55} & \ding{55} & \ding{51} \\
    DCM~\citep{lv2025dcm} & \ding{55} & \ding{55} & \ding{51} \\
    MagicDistillation~\citep{shao2025magicdistillation} & \ding{55} & \ding{55} & \ding{51} \\
    Self-Forcing~\citep{huang2025self} & \ding{55} & \ding{55} & \ding{51} \\
    Causal Forcing~\citep{zhu2026causal} & \ding{55} & \ding{55} & \ding{51} \\
    TMD~\citep{nie2026transition} & \ding{55} & \ding{55} & \ding{51} \\
    rCM~\citep{zheng2025large} & \ding{55} & \ding{55} & \ding{51} \\
    AccVideo~\citep{zhang2025accvideo} & \ding{55} & \ding{55} & \ding{51} \\
    Standard DMD~\citep{yin2024improved} & \ding{55} & \ding{55} & \ding{51} \\
    LCM~\citep{luo2023latent} & \ding{55} & \ding{55} & \ding{51} \\
    sCM~\citep{lu2024simplifying} & \ding{55} & \ding{55} & \ding{51} \\
    AVD~\citep{you2026adaptivevideo} & \ding{55} & \ding{55} & \ding{51} \\
    ASD~\citep{yang2025towards} & \ding{55} & \ding{55} & \ding{51} \\
    LCM (SD) & \ding{51} & \ding{55} & \ding{55} \\
    DMD (SD) & \ding{51} & \ding{55} & \ding{55} \\
    LCM (CFG-E+SD) & \ding{51} & \ding{55} & \ding{51} \\
    DMD (CFG-E+SD) & \ding{51} & \ding{55} & \ding{51} \\
    PCM~\citep{wang2024phased} & \ding{51} & \ding{51} & \ding{55} \\
    \textbf{PP-CFG} & \ding{51} & \ding{51} & \ding{51} \\
    \bottomrule
  \end{tabular}
\end{table}
\FloatBarrier

\section{Limitations}
Experiments on Wan and Qwen-Image support PP-CFG's generality across cross-attention and joint attention, video and image generation, and LCM and DMD distillation. This evidence comes from the evaluated model and training combinations. Extending the same design to additional model families and generation tasks is a natural direction for future work.

\end{document}